\documentclass[runningheads]{llncs}
\usepackage[T1]{fontenc}

\usepackage{graphicx}
\usepackage{amsmath}
\usepackage{amsfonts}
\usepackage{xcolor}
\def\p{{\mathrm{p}}}
\def\q{{\mathrm{q}}}

\def\bx{{\mathbf x}}

\def\bz{{\mathbf z}}

\def\bI{{\mathbf I}}

\def\bX{{\mathbf X}}

\usepackage{subcaption}
\usepackage{cite}

\begin{document}
\title{Evaluating the fairness of generative models: a study on the racial bias of VAEs in dermatological images}
\title{Are generative models fair? A study of racial bias in dermatological image generation}
\titlerunning{Are generative models fair?}

\author{Miguel López-Pérez\inst{1}\and
Søren Hauberg\inst{2} \and
Aasa Feragen\inst{2}}
\authorrunning{M. López-Pérez et al.}
\institute{Instituto Universitario de Investigación en Tecnología Centrada en el Ser Humano,
Universitat Politècnica de València, Spain \and
Technical University of Denmark, Kongens Lyngby, Denmark}
\maketitle              % typeset the header of the contribution
\begin{abstract}
Racial bias in medicine, such as in dermatology, presents significant ethical and clinical challenges. This is likely to happen because there is a significant underrepresentation of darker skin tones in training datasets for machine learning models. While efforts to address bias in dermatology have focused on improving dataset diversity and mitigating disparities in discriminative models, the impact of racial bias on generative models remains underexplored. Generative models, such as Variational Autoencoders (VAEs), are increasingly used in healthcare applications, yet their fairness across diverse skin tones is currently not well understood.
In this study, we evaluate the fairness of generative models in clinical dermatology with respect to racial bias. For this purpose, we first train a VAE with a perceptual loss to generate and reconstruct high-quality skin images across different skin tones. We utilize the Fitzpatrick17k dataset to examine how racial bias influences the representation and performance of these models. Our findings indicate that VAE performance is, as expected, influenced by representation, i.e.~increased skin tone representation comes with increased performance on the given skin tone. However, we also observe, even independently of representation, that the VAE performs better for lighter skin tones. Additionally, the uncertainty estimates produced by the VAE are ineffective in assessing the model's fairness. These results highlight the need for more representative dermatological datasets, but also a need for better understanding the sources of bias in such model, as well as improved uncertainty quantification mechanisms to detect and address racial bias in generative models for trustworthy healthcare technologies. 
\end{abstract}
\section{Introduction}
% Bias in medicine
Racial bias in medicine has been widely documented \cite{williams2015racial,dehon2017systematic}.
This bias is often inherited, or even amplified, by deep learning methods: First, lack of representation can lead to loss of performance and overfitting for underrepresented groups. It has been shown across tasks and datasets in medical imaging that low subgroup representation can be associated with low subgroup performance~\cite{larrazabal2020gender}. Low representation does not, however, always lead to bias~\cite{petersen2022feature}, nor is it the only mechanism leading to bias: Differences in data quality, systematic label errors such as group-dependent over- or underdiagnosis across groups, or group-dependent prediction difficulty can also lead to biased models~\cite{zhou2021radfusion,gichoya2022ai,petersen2023path}. Such limitations can hinder the performance of machine learning algorithms in specific subpopulations, particularly among underrepresented groups. Deploying AI-driven tools in healthcare that may exhibit detrimental performance or consequences for these subgroups is not only unfair but also poses serious risks. For example, an established system that achieves high accuracy may experience an unexpected and dramatic drop in performance when deployed and tested on a different subgroup of the population. Therefore, it is crucial to detect and mitigate these biases to ensure ethical and trustworthy AI in medicine \cite{li2023trustworthy}.

\textbf{In this paper, we contribute} a study of how under- and overrepresentation affects the performance of generative models, exemplified by a VAE. We show that subgroup underrepresentation is associated with subgroup underperformance, but we also obtain a general lower performance on darker skin tones. Finally, we show that the inherent uncertainty quantification built into the VAE is ineffective at capturing underrepresentation and its associated loss of performance.

\section{Related work}

% Bias in dermatology
In dermatology, racial bias is well known, and often attributed to a significant underrepresentation of darker skin tones in clinical data~\cite{benmalek2024impact,kalb2023revisiting,10254437}. Despite promising results in machine learning to accurately classify skin conditions, the existence of racial bias against darker skin tones has been reported \cite{daneshjou2022disparities,10254437}. 
This racial bias may also exacerbate accuracy disparities between light and dark skin tones among non-specialists when machine learning is deployed in the clinical practice to assist physicians  \cite{groh2024deep}. Addressing this issue is essential to develop and deploy safe tools in clinical practice.

Some efforts to overcome racial bias have focused on creating large-scale dermatology image datasets that include metadata on skin tone, typically measured using the Fitzpatrick Skin Type (FST) scale. This scale goes between 1 to 6 (from the lightest to the darkest).
The reference dataset to assess racial bias in dermatological images is the Fitzpatrick17k dataset, which encompasses a wide range of Fitzpatrick Skin Types (FST 1–6) \cite{groh2021evaluating}. While the dataset includes both skin condition and skin type labels, it remains imbalanced, with darker skin tones (FST 4–6) being underrepresented (see Fig.~\ref{fig:fitz17k}). The same study identified that existing dermatology AI models exhibit significant biases, particularly underperforming on darker skin tones. Similar observations were made on the Diverse Dermatology Images dataset~\cite{daneshjou2022disparities}.
Another recent example is the PASSION dataset \cite{gottfrois2024passion}, which focuses on individuals from Sub-Saharan countries and includes FST 3–6, the most common skin type in this region. This dataset aims to address the limitations of previous datasets, which have predominantly focused on lighter skin tones.

\subsection{Generative models and algorithmic bias} 
\begin{figure}[h!]
    \centering
    \includegraphics[width=0.95\linewidth]{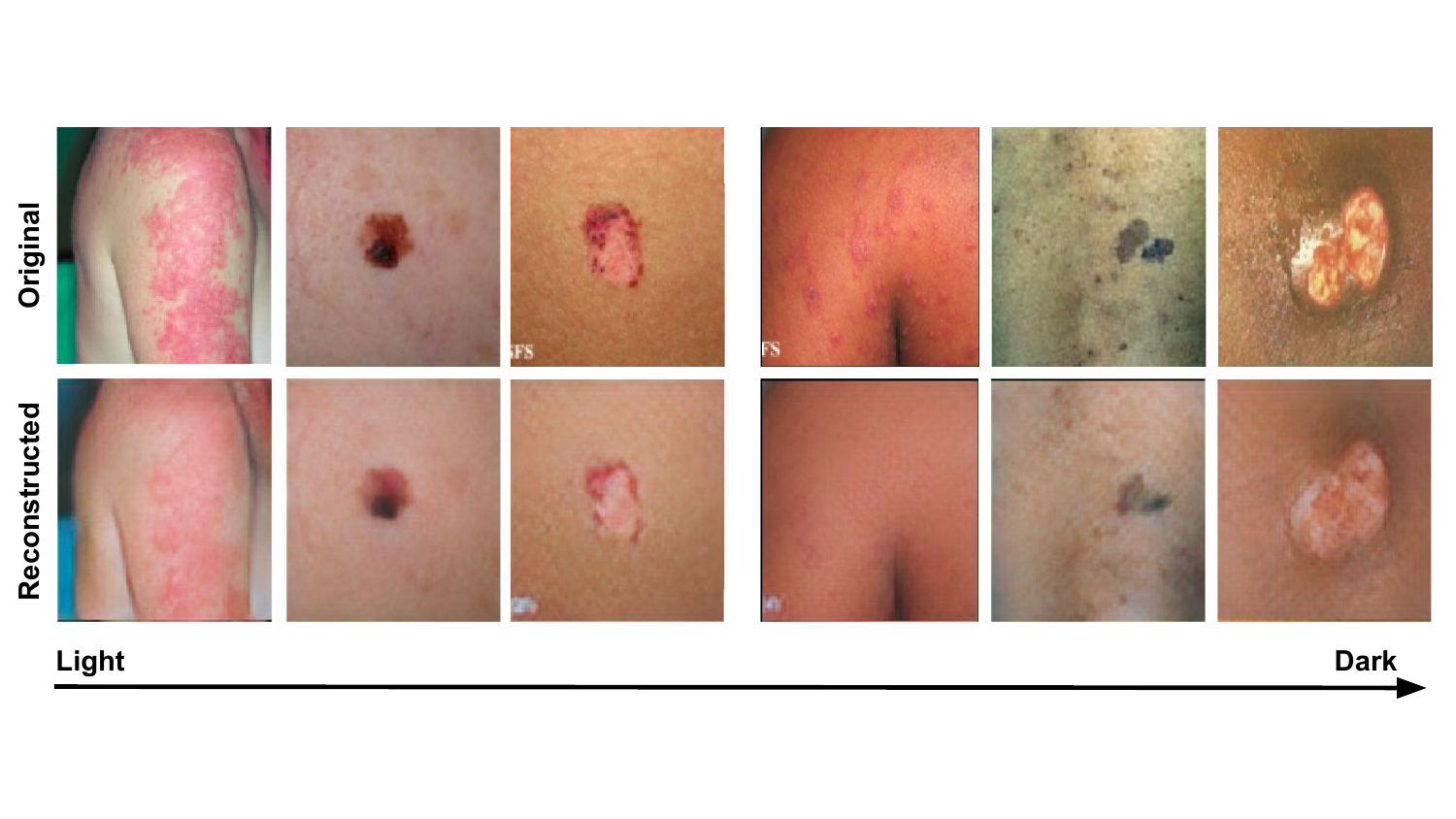}
    \caption{Example predictions from our VAE model trained on a balanced subset of the Fitzpatrick17k dataset, comprising 50/50 light and dark skin tones. Notably, predictions for lighter tones are more accurate and better preserve the lesion compared to those for darker tones.}
    \label{fig:motivation}
\end{figure}

A different attempt to mitigate racial bias was to generate new synthetic samples using large generative models based on diffusion (DALL·E 2) \cite{sagers2022improving}. Generative models are a type of machine learning model that aims to learn the underlying structure of data, allowing them to produce new data points from this distribution. Here, the authors achieved increased accuracy for underrepresented groups by using generative models to balance datasets. However, it remains unclear how racial bias may propagate through generative models. 
For instance, generative models are able to introduce racial bias even when trained on a dataset with balanced representation, as illustrated qualitatively in Figure~\ref{fig:motivation}.

The focus of previous works has been on debiasing discriminative models for the classification of skin conditions. In this paper, we take a step back to address the following questions: 

\begin{itemize}
    \item \textit{How does the performance of a deep generative model, trained to generate clinical images of dermatology, vary across different skin tones?
    \item How does this variation depend on skin tone representation?
    \item How does the inherent uncertainty quantification of the generative model perform in detecting lower subgroup representation, which is associated with lower subgroup performance?}
\end{itemize} 
The way generative models encode and represent data is crucial for understanding their behavior in the context of racial bias. Although these models can generate new synthetic examples, further study is needed to determine whether they reproduce existing biases in this context. In this study, we explore the racial bias of generative models in clinical dermatology images. We utilize a Variational Autoencoder (VAE)~\cite{kingmaauto,rezende14}, a popular generative model with an autoencoder architecture that comes with a probability distribution in the latent space. We use the VAE to generate and reconstruct skin images across different skin types. For this purpose, we employ the Fitzpatrick17k dataset, which offers a varied FST scale, enabling us to investigate how racial bias may affect the performance of VAEs. Finally, we propose new directions toward fair generative models.

\section{Methods}

\subsection{Deep Learning model } 

VAEs \cite{kingmaauto,rezende14} are a popular class of deep generative models. They aim to learn the distribution of a given set of images $\bX=\{\bx\}$ and generate new samples from this distribution $\{\bx_*\}$. It leverages an autoencoder-like architecture combining an \textit{stochastic encoder} $\q_{\phi}(\bx|\bz)$ (parameterized by $\phi$), and a \textit{stochastic decoder} $\p_\theta(\bx|\bz)$ (parameterized by $\theta$). 
The word \textit{stochastic} here means that the encoder and decoder are probability distributions (in contrast to deterministic autoencoders). 

The encoder aims to estimate the optimal parameters of the distribution of the latent representation $\bz$ given an image $\bx$. Here, we use a Gaussian distribution for this, $\q_{\phi}(\bz|\bx)=\mathcal{N}(\mu_{\phi}(\bx), \text{diag}(\sigma_\phi^2(\bx))$. The decoder aims to reconstruct this image from the stochastic representation $\bz$. Here, we utilize the following likelihood model for the decoder, $\p_\theta(\bx|\bz) = \mathcal{N}(\bx|f_\theta(\bz), \bI)$. 
Furthermore, we impose a a prior distribution to the latent variable $\p(\bz)=\mathcal{N}(\bz|0,\bI)$. We perform variational inference to obtain the optimal variational parameters, i.e., $\{\theta,\phi\}$. For this purpose, we maximize a lower bound on the marginal log-likelihood (ELBO):
\begin{equation} \label{eq:elbo}
    \textrm{ELBO}(\theta,\phi;\mathbf{x}) = \mathbb{E}_{q_\phi(\mathbf{z}|\mathbf{x})}\left[\log p_\theta(\mathbf{x}|\mathbf{z}) \right] - \textrm{KL}(q_\phi(\mathbf{z}|\bx)||p(\mathbf{z})),
\end{equation}
where  $\textrm{KL}(q_\phi(\mathbf{z}|\bx)||p(\mathbf{z}))$ is the Kullback-Leibler (KL) divergence between the approximate posterior $q_\phi(\mathbf{z}|\bx)$ and the prior $p(\mathbf{z})$. The first term in Eq.~\ref{eq:elbo} corresponds to the negative of the reconstruction error. In this case, it is the mean square error (MSE) because of the Gaussian likelihood. The second term uses KL divergence to encourage fidelity to the prior distribution. This term can be seen as a regularizer.
The training objective is to maximize the ELBO given the observational data, and the model can be optimized with respect to the variational $\phi$, and generative parameters $\theta$. To estimate the distribution $\p_\theta(\bz|\bx)$, we utilize Monte Carlo sampling with the reparameterization trick to obtain samples of the latent variable $\bz$, i.e., $\bz=\mu(\bx) + \epsilon \odot \sigma(\bx)$, $\epsilon\sim\mathcal{N}(0,\bI)$.

\subsection{Perceptual loss}

When working with high-fidelity images, using a standard VAE may not be the best approach. The generated images are prone to be blurry \cite{huang2018introvae}. This is because using a pixel-by-pixel loss that does not capture spatial correlations between two images. A typical example is the same image translated by a few pixels. To the human eyes, they are practically the same image. However, it will result in high pixel-by-pixel loss.

This drawback of VAEs working with images has been solved by feature perceptual loss \cite{hou2017deep}. This loss between two images is defined by the difference between the hidden maps in a pretrained convolutional neural network $\Phi$. The main idea behind this approach is that the feature maps of this network have sensible information about spatial correlation in the images. Namely, it will approach better to the human vision. As a result, we can get a better visual quality output image. The loss function of the VAE with the perceptual loss used in this work is as follows,

 \begin{equation}\label{eq:perploss}
     \mathcal{L} = \mathrm{ELBO} + \frac{1}{2C^lW^lH^l}\sum^{C^l}_{c=1}\sum^{W^l}_{w=1}\sum^{H^l}_{h=1}(\Phi(\bx)^l_{c,w,h} - \Phi(f_\theta(x))_{c,w,h})^2,
 \end{equation}
 where $C^l, H^l, W^l$ are the channels, height, and widht of the $l$-th feature map of the network, respectively.

\subsection{Fitzpatrick17k dataset}

For our study, we utilize the Fitzpatrick17k dataset \cite{groh2021evaluating}, which is a large publicly available dataset with associated information on skin tone represented using the Fitzpatrick scale. This dataset was sourced from two open-source dermatology atlases, and  comprises 16,577 images with corresponding skin condition and Fitzpatrick labels. The skin conditions have associated labels
at different levels of granularity. The fine-grained labels for specific conditions
are grouped into 9 superclasses, which are further consolidated into 3 super-
classes. The Fitzpatrick scale ranges from 1 to 7, with -1 representing missing values.

\begin{figure}[h!]
    \centering
    \includegraphics[width=0.95\linewidth]{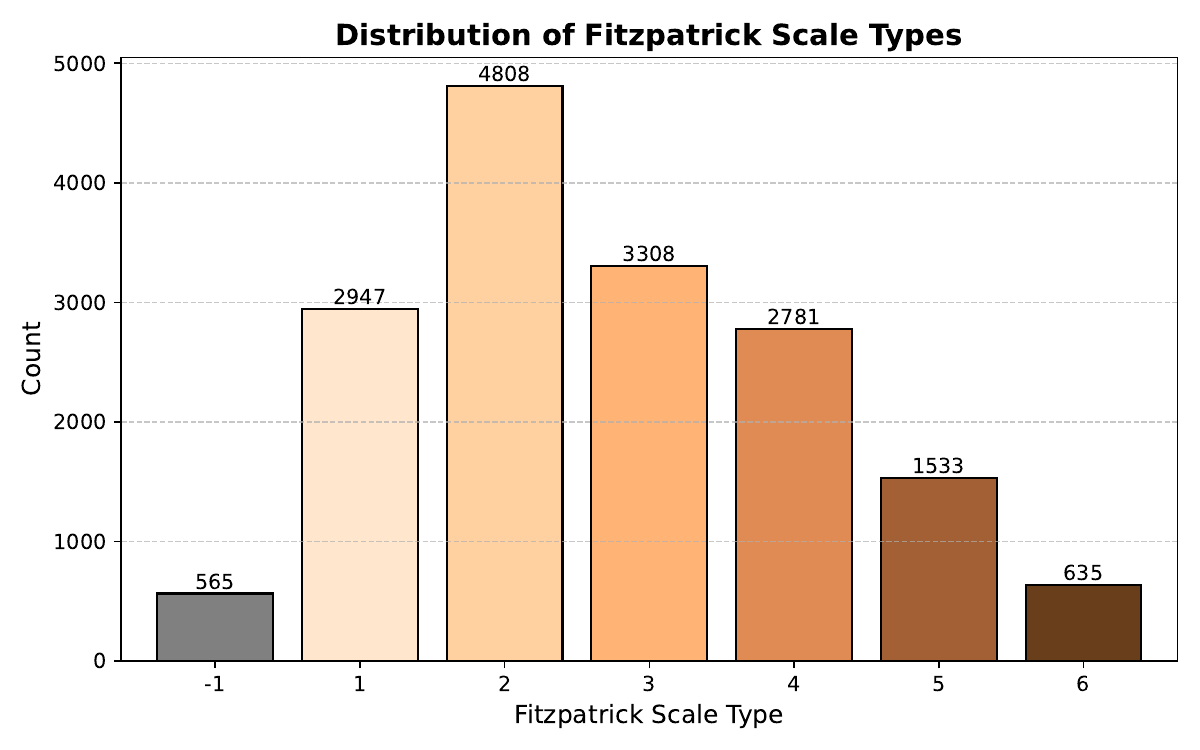}
    \caption{Distribution of samples according to the FST in the Fitzpatrick17k dataset \cite{groh2021evaluating}. The label `-1' represents missing values.
    %For this study, we discarded -1 (missing values) and FST 3-4 to avoid ambiguities.
    }
    \label{fig:fitz17k}
\end{figure}

The Fitzpatrick skin-type labels were assigned by a team of human annotators from Scale AI. The labels were obtained by a consensus process from two to five annotators. They carried out a process to achieve the desired level of agreement, resulting in 72,277 annotations in total. The distribution of the Fitzpatrick labels of this dataset is depicted in Figure~\ref{fig:fitz17k}. 
We observe that darker skin tones are notably underrepresented also in this dataset, and we will need to subsample to create balanced scenarios. %This information is particularly interesting for us since we focus on assessing the racial bias of the generative model.

\subsection{Experimental details}

\paragraph{Neural network architecture.} The encoder and decoder of the VAE utilize residual blocks along with down-sampling and up-sampling, respectively. Each convolution block employs batch normalization and ELU activation for stable training.  The resulting latent space has shape $8\times8\times64$. For the perceptual loss in equation \eqref{eq:perploss}, we used a VGG19 pretrained on ImageNet \cite{simonyan2015deepconvolutionalnetworkslargescale}. The perceptual loss was computed using the output of every convolutional layer, $16$ layers in this case.

\paragraph{Optimization details.} We train the models for $15$ epochs after which we observed a convergence of the training loss. The optimizer was Adam with a learning rate of $10^{-4}$. The training was performed in mini-batches of $64$ samples.

\paragraph{Implementation.} We utilize the implementation of a VAE with perceptual loss available at GitHub\footnote{\url{https://github.com/LukeDitria/CNN-VAE} (accessed 2024-02-17).}. The code was implemented in PyTorch 2.0.1. The code was executed using one GPU NVIDIA GeForce RTX 3090 to train and evaluate the models.

\paragraph{Sensitive groups.} We divided the Fitzpatrick17k into two different groups. One with lighter skin, comprising FST 1-2, and one with darker skin, comprising FST 4-6. Images without a Fitzpatrick label were discarded, along with those labeled as FST 3-4, to avoid ambiguities, as these categories lie on the boundary between lighter and darker skin tones.

\paragraph{Experimental setting.} We imitate the experimental design from~\cite{larrazabal2020gender}, originally created to showcase how subgroup performance depends on subgroup representation for image classification tasks. The experiment is repeated ten independent times. For each run, two test sets of 500 samples each are sampled from the dataset, one with lighter images (FST 1-2) and the other with darker ones (FST 5-6). Then, with the remaining images, we independently sample three training set configurations (with replacement) of $1668$ samples. The `Dataset A -- Light' has 100\% of lighter images, the `Dataset B -- Mixed' set has 50/50, and the `Dataset C -- Black' set has 100\% of darker images. 

\paragraph{Metrics. } To monitor how the VAE performance on subgroups depends on representation, we study two different metrics across the different test sets and training runs: To monitor reconstruction performance, we utilize the likelihood, i.e., the MSE or reconstruction error. To monitor the inherent uncertainty quantified by the model, we measure the latent standard deviation given by the stochastic encoder.

\section{Results}

\begin{figure}[t]
    \centering
    \includegraphics[width=0.95\linewidth]{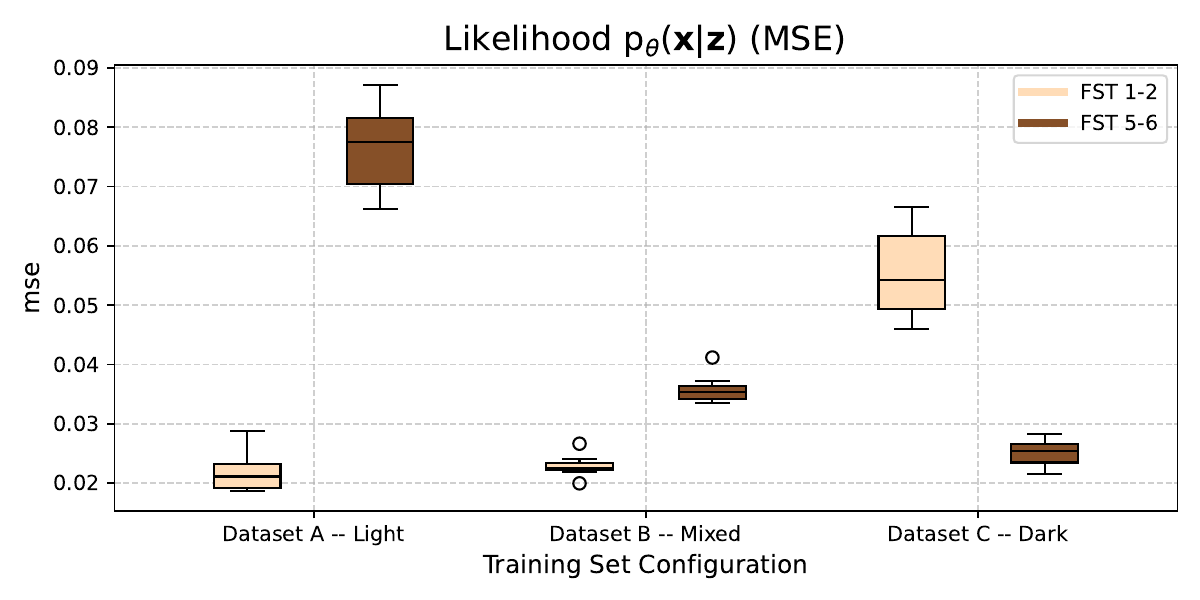}
    \caption{Likelihood or Mean Square Error (MSE) of the VAEs in the test sets. }
    \label{fig:mse}
\end{figure}

Figure~\ref{fig:mse} shows boxplots of the likelihood (or MSE) for our VAE with perceptual loss across the three different training set configurations. For each training set configuration, we plot the results for the both test sets: light skin (FST 1-2) and dark skin (FST 5-6). 
The VAE generally performs well with an MSE below 0.07, which means that the neural network generally assigns a high likelihood to the data. 

\subsection{Subgroup reconstruction performance depends on subgroup representation}

Looking at groupwise MSE, we see the expected trend, where the MSE decreases when the representation increases: When moving towards the right, we increase the representation of dark skin, and the dark skin MSE goes down. When moving towards the left, we increase the representation of light skin, and the light skin MSE goes down.

\subsection{Overall bias against darker skin}
However, the boxplot also shows an overall performance gap between the two population subgroups: The MSE is generally lower for the light skin group. This is noticeable under the `Dataset B -- Mixed', when training with an equal 50/50 sample distribution from both subgroups. Here, a significant performance gap still persists between the two groups. But this is further emphasized when considering the more extreme training scenarios: 
Comparing the performance for dark skin (FST 5-6) using `Dataset A - Light' (only trained with FST 1-2) with the performance for light skin (FST 1-2) using `Dataset C -- Dark' (only trained with FST 5-6), there is a performance gap showing that the worst case performance for the dark skinned group is noticeably lower than the worst case performance for the light skinned group. 

Even for the best case performances, comparing the performance for dark skin (FST 5-6) using `Dataset C -- Dark' (only trained with FST 5-6) with the performance for light skin (FST 1-2) using `Dataset A -- Light' (only trained with FST 1-2), we still see a slightly better performance on the light skinned group.

Overall, our results indicate a general underperformance on the dark skinned group that does not seem to be explained only by lack of representation.

% This finding suggests that the darker skin tones in this dataset make it more difficult to extract meaningful features for the reconstruction than the lighter skin tones. In configuration `C', when training with samples of FST 5-6, the opposite occurs compared to configuration `A'. Here, the performance gap is not as large as in `A'. This corroborates the hypothesis that the generative model is generally biased against darker skin tones. 

\begin{figure}[h!]
    \centering
    \includegraphics[trim={0 3cm 0 0}, clip, width=0.95\linewidth]{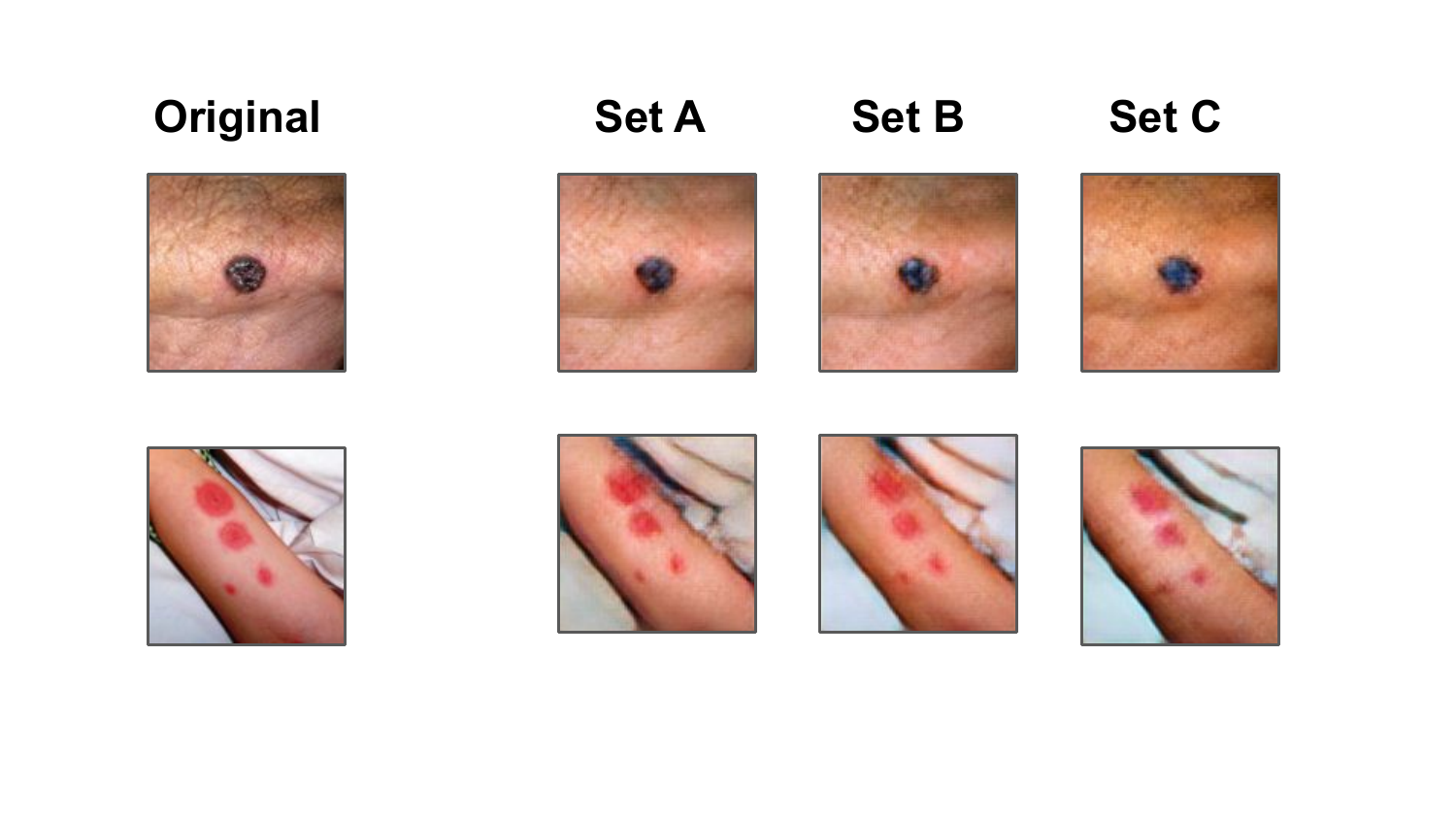}
    \caption{Example reconstruction of lighter skin tones. Reconstructions are produced using three training set configurations: `A', `B', and `C'.}
    \label{fig:whiteplot}
\end{figure}

\begin{figure}[h!]
    \centering
    \includegraphics[trim={0 3cm 0 0}, clip, width=0.95\linewidth]{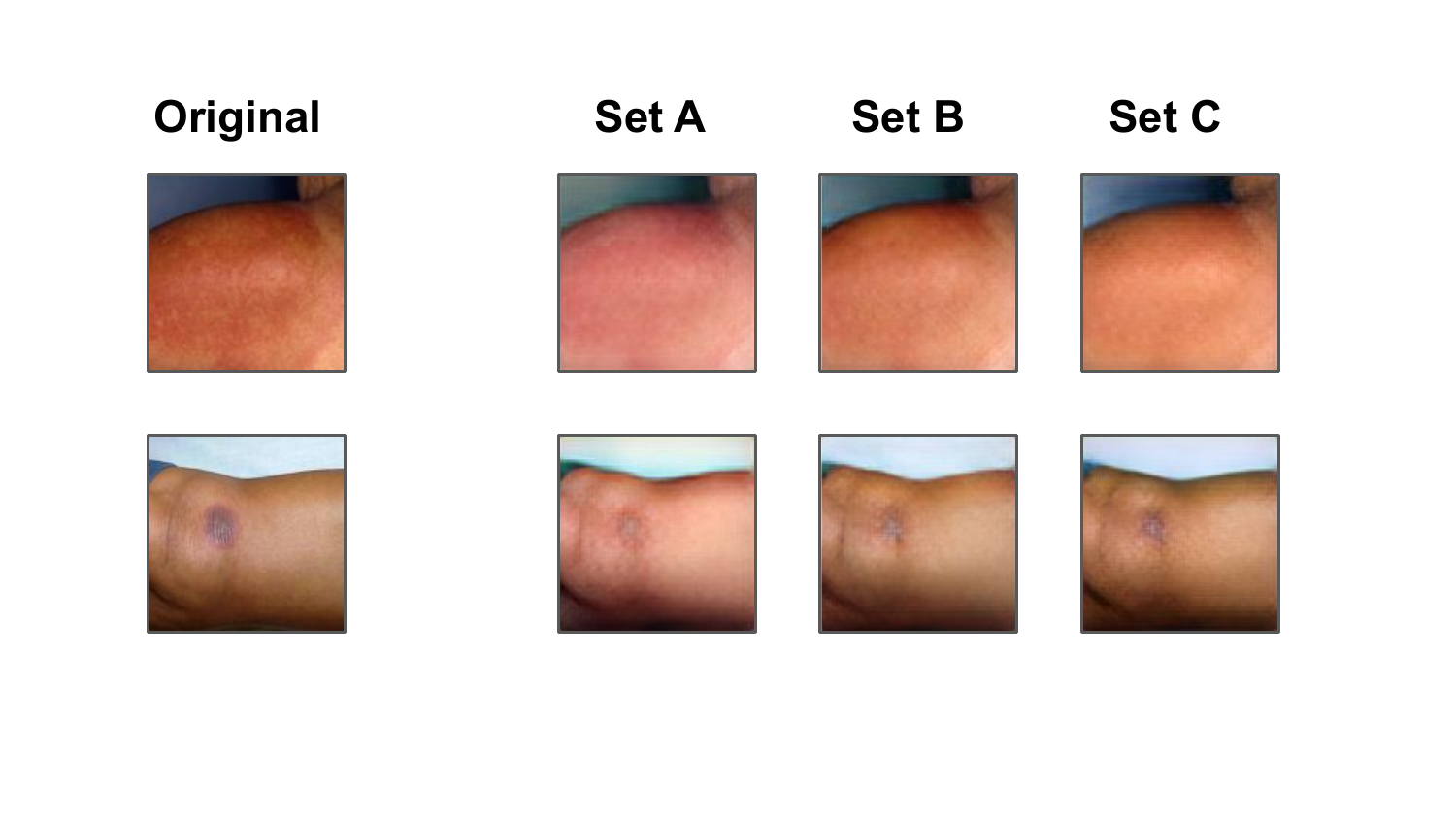}
    \caption{Example reconstruction of darker skin tones. Reconstructions are produced using three training set configurations: `A', `B', and `C'.}
    \label{fig:blackplot}
\end{figure}

To complement these results with qualitative evaluation, we include visualizations of the predicted reconstructions in Figures~\ref{fig:whiteplot} and~\ref{fig:blackplot}. We include reconstructions produced by each of the three different training scenarios to illustrate the effect of the training set representation on subgroup performance. For lighter skin tones (see Fig. \ref{fig:whiteplot}), the reconstructions do not show significant visual variation. However, for darker skin tones (see Fig. \ref{fig:blackplot}), the visibly highest quality reconstructions are obtained with `Dataset C -- Dark' (where 100\% of the samples are dark). Notably, the color tone changes as more dark skin samples are introduced into the training set. These observations support that darker skin tones in clinical skin images are more sensitive to the training set composition, and the recovery of skin tone and details of the injury is more challenging.

\begin{figure}[t]
    \centering
    \includegraphics[width=0.95\linewidth]{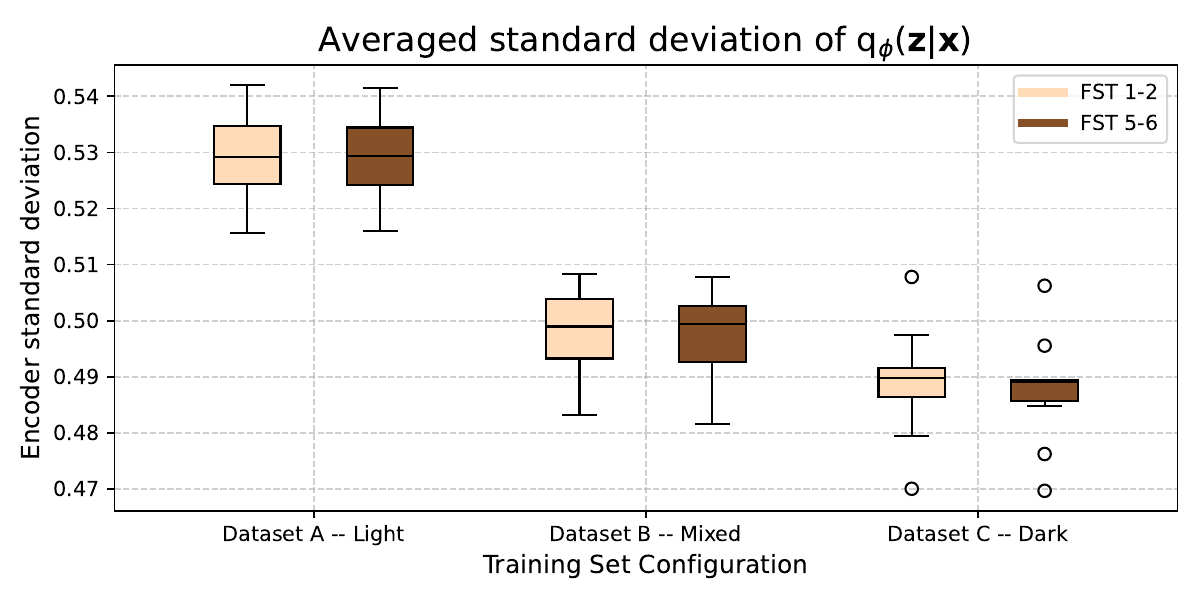}
    \caption{Averaged standard deviation of the latent variable of the VAEs in the test sets.}
    \label{fig:std}
\end{figure}

\subsection{Inherent VAE uncertainty quantification does not serve as a red flag warning against subgroup underrepresentation and underperformance}

Figure~\ref{fig:std} depicts the values of the averaged latent standard deviation of the latent variable $\bz$ for our VAE with perceptual loss in the three different configurations of the training set. This figure has been made using the same process as in the previous Figure~ \ref{fig:mse}. Here, while we observe a reduction in the standard deviation as we include more dark skinned images in the training set, we do not observe any differences between the estimated standard deviation across the two population subgroups at test time. Instead, the standard deviation follows a similar distribution for both test sets. This suggests that the inherent uncertainty quantification mechanism of the VAE to measure the uncertainty of the latent representation is not a proxy for subgroup representation, and therefore is unlikely to be useful to monitor the fairness of the model since the performance, assessed with the likelihood in Figure~\ref{fig:mse}, has a gap between the two test sets which is not replicated in Figure~\ref{fig:std}.

\section{Discussion and conclusion}

Fairness in AI systems for healthcare is urgently needed to ensure that trustworthy AI systems are deployed in clinical practice. Clinical images of dermatology have been observed that show racial bias, as they often fail to adequately represent the wide range of skin tones present in the population. This bias has been widely studied in the context of predictive AI for computer-aided diagnosis. In this study, we show initial results on how this bias may affect deep generative models.

\subsection{Overall findings}

We have shown, as expected, that \textbf{subgroup representation affects VAE reconstruction performance}: We observe that the VAE assigns a lower likelihood to reconstructions of skin tones not seen frequently during training. We do, however, also make some additional unexpected observations:

First, we observe that \textbf{the latent standard deviation of the VAE does not provide any information about the subgroup underrepresentation or bias} in the model, and it can therefore not be used as a proxy to detect fairness problems. This raises concerns about our abilities to discover potential biases when these models are deployed to generate synthetic images of skin pathologies. This leaves an interesting future research question of whether it is possible to develop enhanced uncertainty mechanisms that can effectively capture and relate to the fairness of VAEs. This would allow those uncertainty measures to work as algorithmic bias warning signs.

Second, the VAE model proposed in this study is \textbf{generally biased against darker skin tones}. Even when the training set is balanced with a 50/50 distribution of the two subgroups, the VAE is still biased against darker skin tones. In general, lighter skin tones appear to be easier to reconstruct. This suggests that there is more at play in dermatological AI bias than just a simple representation bias.

%The results obtained here suggest that the predicted uncertainty of VAEs is not adequate or informative for assessing the fairness of deep generative models. 

\subsection{The reason for the overall performance disparity is unclear}

Looking back at the examples from Figures~\ref{fig:whiteplot} and~\ref{fig:blackplot}, one might hypothesize that a potential reason for the overall performance gap between light and dark skinned test sets comes from differences in the subgroups' diseases: The light skinned examples in Figure~\ref{fig:whiteplot} showcase clearly delineated lesions, whereas the dark skinned examples in Figure~\ref{fig:blackplot} shows more diffuse, textured rashes whose images might be more difficult for the model to reconstruct. To assess whether this could explain the performance gap, we therefore investigated the distribution of diseases between the different skin tones.

\begin{figure}[h!]%[!t]
        \centering
        \begin{subfigure}[t]{0.3\textwidth}
            \centering
            \includegraphics[width=\linewidth, keepaspectratio]{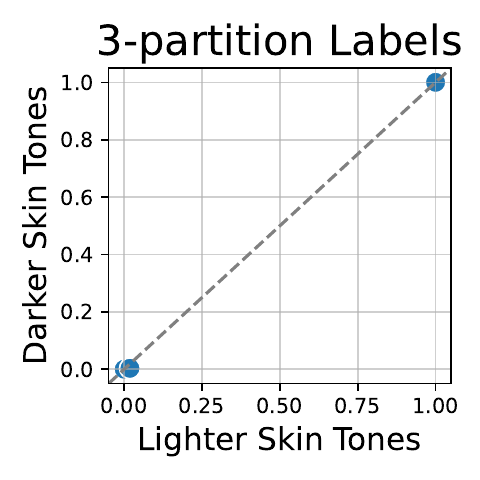}
            \caption{Coarse level of diagnostic granularity.}
            %\label{fig:fig-a}
        \end{subfigure}
        \hfill
        \begin{subfigure}[t]{0.3\textwidth}
            \centering
            \includegraphics[width=\linewidth, keepaspectratio]{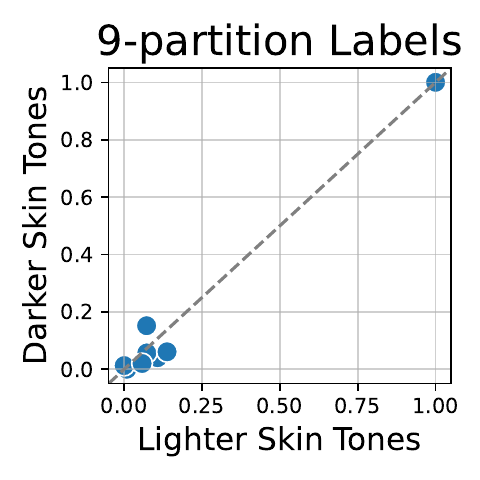}
            \caption{Medium level of diagnostic granularity.}
            %\label{fig:fig-b}
        \end{subfigure}
         \hfill
        \begin{subfigure}[t]{0.3\textwidth}
            \centering
            \includegraphics[width=\linewidth, keepaspectratio]{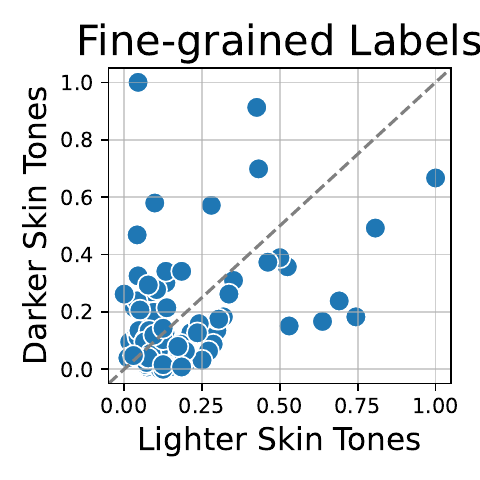}
            \caption{Fine-grained level of diagnostic granularity.}
            %\label{fig:fig-b}
        \end{subfigure}
        \caption{Normalized count of skin conditions categorized across varying levels of diagnostic granularity within the hierarchy of labels.}
        \label{fig:classes}
        \end{figure}

Recall that diagnostic labels were given at three different levels of granularity, ranging from coarse, to middle, to fine-grained. At each level of granularity, we assess whether there are major differences in representation between the two groups. The scatter plot shown in Figure \ref{fig:classes} compares the distribution of condition prevalences for light (x-axis) versus dark (y-axis) skin tone groups, across the diagnostic hierarchy. We observe that at higher levels of the hierarchy, the distribution shows minimal differences across subgroups. However, at the finest-grained level, the distribution of conditions varies substantially between subgroups. 

\begin{figure}%[!t]
        \centering
        \begin{subfigure}[t]{0.3\textwidth}
            \centering
            \includegraphics[width=\linewidth, keepaspectratio]{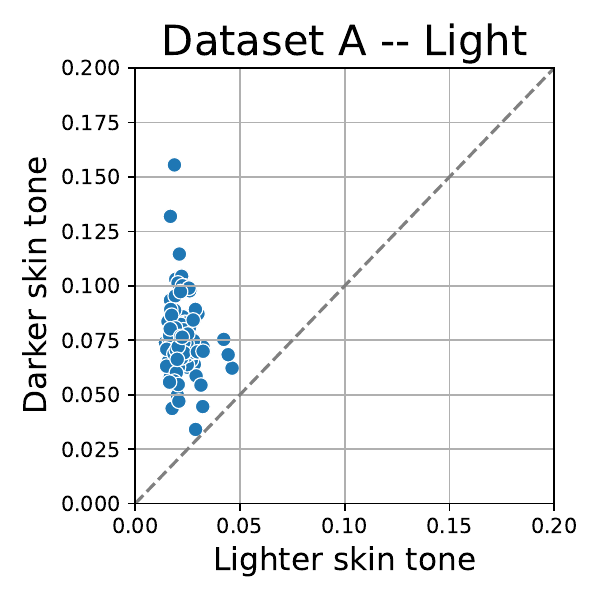}
            \caption{Coarse level of diagnostic granularity.}
            %\label{fig:fig-a}
        \end{subfigure}
        \hfill
        \begin{subfigure}[t]{0.3\textwidth}
            \centering
            \includegraphics[width=\linewidth, keepaspectratio]{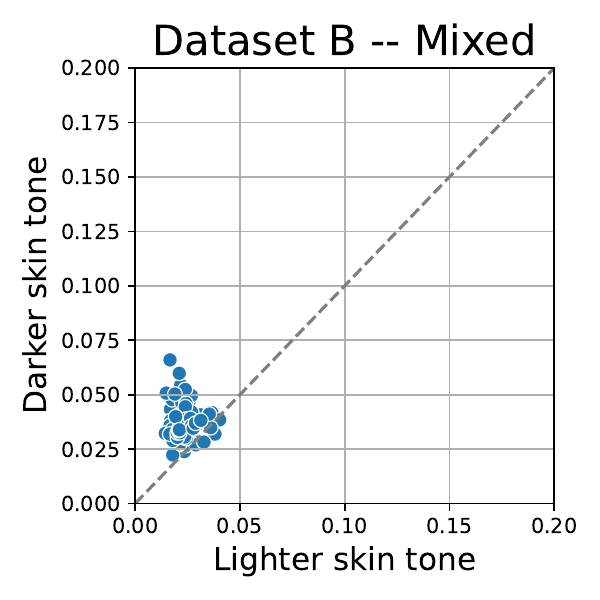}
            \caption{Medium level of diagnostic granularity.}
            %\label{fig:fig-b}
        \end{subfigure}
         \hfill
        \begin{subfigure}[t]{0.3\textwidth}
            \centering
            \includegraphics[width=\linewidth, keepaspectratio]{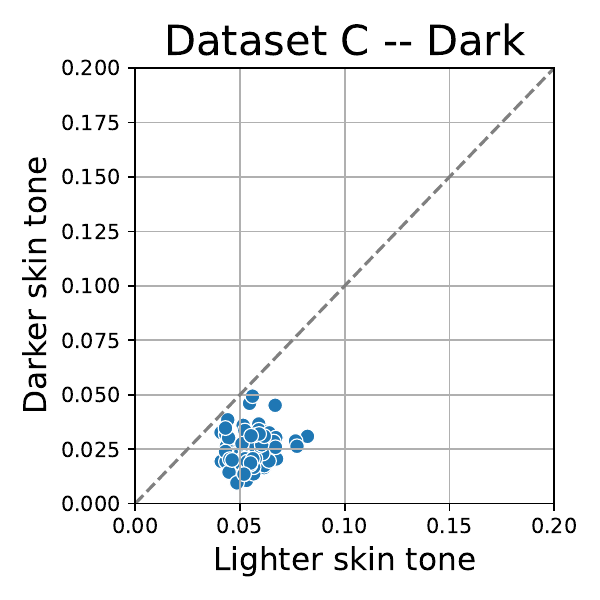}
            \caption{Fine-grained level of diagnostic granularity.}
            %\label{fig:fig-b}
        \end{subfigure}
        \caption{Averaged MSE of skin conditions based on  fine-grained labels.}
        \label{fig:mse_classes}
        \end{figure}

To see how this relates to performance across these conditions, we show the MSE for the fine-grained labels across the three different training scenarios in Figure~\ref{fig:mse_classes}. In general, we observe that the behavior is uneven. `Dataset B -- Mixed' represents the fairest scenario, as the labels are closer to the $y=x$ line. However, performance remains biased against darker skin tones. 

This leaves us with an open question: Can we disentangle the influence of skin tone versus specific skin condition on the results during image generation? 

\subsection{Outlook}

Second, and perhaps even more importantly: How do we assess the quality of datasets for algorithmic fairness analysis? We have already shown that there may be hidden stratifications of our skin tone subgroups from the point of view of diagnoses, which opens further questions about whether these hidden groups can also work as shortcuts for models, further complicating the picture~\cite{oakden2020hidden,olesen2024slicing}. But skin tone is in itself a complex and controversial attribute, and recent work has focused on whether measures such as Fitzpatrick -- which we rely on in this paper -- are even sufficiently correct, or measuring the appropriate features for racial bias~\cite{schumann2024consensus,kalb2023revisiting,kinyanjui2020fairness}.

As a final note: While we have made an effort here to study the effect of how different levels of training set representation affects models, such experiments are impossible within most publicly available dermatological datasets because virtually all images are of light-skinned subjects. As such, we note that `Dataset A -- Light' (composed by only FST 1-2), which demonstrates the highest possible performance disparity in Figure~\ref{fig:mse}, is the typical scenario encountered in real life. This just highlights that there are, indeed, problems yet to be solved, and that we should expect skin tone biases in generative AI models.

%\section{Conclusion}

%
\begin{credits}

\subsubsection{\ackname}
This work was done during Miguel López-Pérez's visit to the Technical University of Denmark in the summer of 2024. This visit was funded by MICIU/AEI/10.13039/501100011033 and the European Union’s “NextGenerationEU”/PRTR through grant JDC2022-048318-I. The work was partly funded by the Novo Nordisk Foundation through the Center for Basic Machine Learning Research in Life Science (NNF20OC0062606). AF was supported by research grant (0087102) from the Novo Nordisk Foundation. SH was supported by a research grant (42062) from VILLUM FONDEN and from the European Research Council (ERC) under the European Union’s Horizon programme (grant agreement 101125993).
\subsubsection{\discintname}
The authors have no competing interests to declare that are
relevant to the content of this article.
\end{credits}
\bibliographystyle{splncs04}
\bibliography{refs}

\begin{thebibliography}{10}
\providecommand{\url}[1]{\texttt{#1}}
\providecommand{\urlprefix}{URL }
\providecommand{\doi}[1]{https://doi.org/#1}

\bibitem{10254437}
Almuzaini, A.A., Dendukuri, S.K., Singh, V.K.: Toward fairness across skin tones in dermatological image processing. In: 2023 IEEE 6th International Conference on Multimedia Information Processing and Retrieval (MIPR). pp.~1--7 (2023). \doi{10.1109/MIPR59079.2023.00030}

\bibitem{benmalek2024impact}
Benmalek, A., Cintas, C., Tadesse, G.A.: Impact of skin tone diversity on out-of-distribution detection methods in dermatology. In: International Conference on Medical Image Computing and Computer-Assisted Intervention (2024)

\bibitem{daneshjou2022disparities}
Daneshjou, R., Vodrahalli, K., Novoa, R.A., Jenkins, M., Liang, W., Rotemberg, V., Ko, J., Swetter, S.M., Bailey, E.E., Gevaert, O., et~al.: Disparities in dermatology ai performance on a diverse, curated clinical image set. Science advances  \textbf{8}(31),  eabq6147 (2022)

\bibitem{dehon2017systematic}
Dehon, E., Weiss, N., Jones, J., Faulconer, W., Hinton, E., Sterling, S.: A systematic review of the impact of physician implicit racial bias on clinical decision making. Academic Emergency Medicine  \textbf{24}(8),  895--904 (2017)

\bibitem{gichoya2022ai}
Gichoya, J.W., Banerjee, I., Bhimireddy, A.R., Burns, J.L., Celi, L.A., Chen, L.C., Correa, R., Dullerud, N., Ghassemi, M., Huang, S.C., et~al.: Ai recognition of patient race in medical imaging: a modelling study. The Lancet Digital Health  \textbf{4}(6),  e406--e414 (2022)

\bibitem{gottfrois2024passion}
Gottfrois, P., Gr{\"o}ger, F., Andriambololoniaina, F.H., Amruthalingam, L., Gonzalez-Jimenez, A., Hsu, C., Kessy, A., Lionetti, S., Mavura, D., Ng’ambi, W., et~al.: Passion for dermatology: Bridging the diversity gap with pigmented skin images from sub-saharan africa. In: International Conference on Medical Image Computing and Computer-Assisted Intervention. pp. 703--712. Springer (2024)

\bibitem{groh2024deep}
Groh, M., Badri, O., Daneshjou, R., Koochek, A., Harris, C., Soenksen, L.R., Doraiswamy, P.M., Picard, R.: Deep learning-aided decision support for diagnosis of skin disease across skin tones. Nature Medicine  \textbf{30}(2),  573--583 (2024)

\bibitem{groh2021evaluating}
Groh, M., Harris, C., Soenksen, L., Lau, F., Han, R., Kim, A., Koochek, A., Badri, O.: Evaluating deep neural networks trained on clinical images in dermatology with the fitzpatrick 17k dataset. In: Proceedings of the IEEE/CVF Conference on Computer Vision and Pattern Recognition. pp. 1820--1828 (2021)

\bibitem{hou2017deep}
Hou, X., Shen, L., Sun, K., Qiu, G.: Deep feature consistent variational autoencoder. In: 2017 IEEE winter conference on applications of computer vision (WACV). pp. 1133--1141. IEEE (2017)

\bibitem{huang2018introvae}
Huang, H., He, R., Sun, Z., Tan, T., et~al.: Introvae: Introspective variational autoencoders for photographic image synthesis. Advances in neural information processing systems  \textbf{31} (2018)

\bibitem{kalb2023revisiting}
Kalb, T., Kushibar, K., Cintas, C., Lekadir, K., Diaz, O., Osuala, R.: Revisiting skin tone fairness in dermatological lesion classification. In: Workshop on Clinical Image-Based Procedures. pp. 246--255. Springer (2023)

\bibitem{kingmaauto}
Kingma, D.P., Welling, M.: Auto-encoding variational bayes. In: International Conference on Learning Representations (ICLR) (2014)

\bibitem{kinyanjui2020fairness}
Kinyanjui, N.M., Odonga, T., Cintas, C., Codella, N.C., Panda, R., Sattigeri, P., Varshney, K.R.: Fairness of classifiers across skin tones in dermatology. In: International Conference on Medical Image Computing and Computer-Assisted Intervention. pp. 320--329. Springer (2020)

\bibitem{larrazabal2020gender}
Larrazabal, A.J., Nieto, N., Peterson, V., Milone, D.H., Ferrante, E.: Gender imbalance in medical imaging datasets produces biased classifiers for computer-aided diagnosis. Proceedings of the National Academy of Sciences  \textbf{117}(23),  12592--12594 (2020)

\bibitem{li2023trustworthy}
Li, B., Qi, P., Liu, B., Di, S., Liu, J., Pei, J., Yi, J., Zhou, B.: Trustworthy ai: From principles to practices. ACM Computing Surveys  \textbf{55}(9),  1--46 (2023)

\bibitem{oakden2020hidden}
Oakden-Rayner, L., Dunnmon, J., Carneiro, G., R{\'e}, C.: Hidden stratification causes clinically meaningful failures in machine learning for medical imaging. In: Proceedings of the ACM conference on health, inference, and learning. pp. 151--159 (2020)

\bibitem{olesen2024slicing}
Olesen, V., Weng, N., Feragen, A., Petersen, E.: Slicing through bias: Explaining performance gaps in medical image analysis using slice discovery methods. In: MICCAI Workshop on Fairness of AI in Medical Imaging. pp. 3--13. Springer (2024)

\bibitem{petersen2022feature}
Petersen, E., Feragen, A., da~Costa~Zemsch, M.L., Henriksen, A., Wiese~Christensen, O.E., Ganz, M., Initiative, A.D.N.: Feature robustness and sex differences in medical imaging: a case study in mri-based alzheimer’s disease detection. In: International Conference on Medical Image Computing and Computer-Assisted Intervention. pp. 88--98. Springer (2022)

\bibitem{petersen2023path}
Petersen, E., Holm, S., Ganz, M., Feragen, A.: The path toward equal performance in medical machine learning. Patterns  \textbf{4}(7) (2023)

\bibitem{rezende14}
Rezende, D.J., Mohamed, S., Wierstra, D.: Stochastic backpropagation and approximate inference in deep generative models. In: Xing, E.P., Jebara, T. (eds.) Proceedings of the 31st International Conference on Machine Learning. Proceedings of Machine Learning Research, vol.~32, pp. 1278--1286. PMLR, Bejing, China (22--24 Jun 2014)

\bibitem{sagers2022improving}
Sagers, L.W., Diao, J.A., Groh, M., Rajpurkar, P., Adamson, A., Manrai, A.K.: Improving dermatology classifiers across populations using images generated by large diffusion models. In: NeurIPS 2022 Workshop on Synthetic Data for Empowering ML Research (2022)

\bibitem{schumann2024consensus}
Schumann, C., Olanubi, F., Wright, A., Monk, E., Heldreth, C., Ricco, S.: Consensus and subjectivity of skin tone annotation for ml fairness. Advances in Neural Information Processing Systems  \textbf{36} (2024)

\bibitem{simonyan2015deepconvolutionalnetworkslargescale}
Simonyan, K., Zisserman, A.: Very deep convolutional networks for large-scale image recognition (2015), \url{https://arxiv.org/abs/1409.1556}

\bibitem{williams2015racial}
Williams, D.R., Wyatt, R.: Racial bias in health care and health: challenges and opportunities. Jama  \textbf{314}(6),  555--556 (2015)

\bibitem{zhou2021radfusion}
Zhou, Y., Huang, S.C., Fries, J.A., Youssef, A., Amrhein, T.J., Chang, M., Banerjee, I., Rubin, D., Xing, L., Shah, N., et~al.: Radfusion: Benchmarking performance and fairness for multimodal pulmonary embolism detection from ct and ehr. arXiv preprint arXiv:2111.11665  (2021)

\end{thebibliography}

\end{document}